\documentclass{article}
\usepackage{spconf,amsmath,graphicx}
\usepackage{graphicx}
\usepackage{amsmath}
\usepackage{amssymb}
\usepackage{booktabs}
\usepackage{multirow}
\usepackage{empheq}
\usepackage{subcaption}
\usepackage{arydshln}
\usepackage{lipsum}  

\usepackage{pifont}

\usepackage[pagebackref,breaklinks,colorlinks]{hyperref}

\usepackage[capitalize]{cleveref}
\crefname{section}{Sec.}{Secs.}
\Crefname{section}{Section}{Sections}
\Crefname{table}{Table}{Tables}
\crefname{table}{Tab.}{Tabs.}

\title{RESSCAL3D: Resolution Scalable 3D Semantic Segmentation of Point Clouds}
\name{Remco Royen$^{\star \dagger}$, Adrian Munteanu$^{\star \dagger}$\thanks{This work is funded by Fonds Wetenschappelijk Onderzoek (FWO) - 1S89420N and Innoviris within the research project SPECTRE.\newline © 2023 IEEE. Personal use of this material is permitted. Permission from IEEE must be obtained for all other uses, in any current or future media, including reprinting/republishing this material for advertising or promotional purposes, creating new collective works, for resale or redistribution to servers or lists, or reuse of any copyrighted component of this work in other works. DOI: 10.1109/ICIP49359.2023.10222338}}
\address{$^{\star}$Department ETRO, Vrije Universiteit Brussel, Pleinlaan 2, B-1050 Brussels, Belgium\\
$^{\dagger}$imec, Kapeldreef 75, B-3001 Leuven, Belgium\\
Email: \{\textit{remco.royen, adrian.munteanu}\}@vub.be}

\begin{document}
%\ninept
%
\maketitle
\begin{abstract}
% The abstract should appear at the top of the left-hand column of text, about
% 0.5 inch (12 mm) below the title area and no more than 3.125 inches (80 mm) in
% length.  Leave a 0.5 inch (12 mm) space between the end of the abstract and the
% beginning of the main text.  The abstract should contain about 100 to 150
% words, and should be identical to the abstract text submitted electronically
% along with the paper cover sheet.  All manuscripts must be in English, printed
% in black ink.

While deep learning-based methods have demonstrated outstanding results in numerous domains, some important functionalities are missing. Resolution scalability is one of them. In this work, we introduce a novel architecture, dubbed RESSCAL3D, providing resolution-scalable 3D semantic segmentation of point clouds. In contrast to existing works, the proposed method does not require the whole point cloud to be available to start inference. Once a low-resolution version of the input point cloud is available, first semantic predictions can be generated in an extremely fast manner. This enables early decision-making in subsequent processing steps. As additional points become available, these are processed in parallel. To improve performance, features from previously computed scales are employed as prior knowledge at the current scale. Our experiments show that RESSCAL3D is 31-62\% faster than the non-scalable baseline while keeping a limited impact on performance. To the best of our knowledge, the proposed method is the first to propose a resolution-scalable approach for 3D semantic segmentation of point clouds based on deep learning.

% Deep learning-based methods have shown to achieve excellent results in a variety of domains, however, some important assets are absent. Quality scalability is one of them. In this work, we introduce a novel and generic neural network layer, named MaskLayer. It can be integrated in any feedforward network, allowing quality scalability by design by creating embedded feature sets. These are obtained by imposing a specific structure of the feature vector during training. To further improve the performance, a masked optimizer and a balancing gradient rescaling approach are proposed. Our experiments show that the cost of introducing scalability using MaskLayer remains limited. In order to prove its generality and applicability, we integrated the proposed techniques in existing, non-scalable networks for point cloud compression and semantic hashing with excellent results. To the best of our knowledge, this is the first work presenting a generic solution able to achieve quality scalable results within the deep learning framework.

\end{abstract}
%

% QUESTIONS:
% 
% Are there too much self-citations?
% Maybe add Baseline in last column visuals?
% \textcolor{red}{Depending on the exact application, the second scale can be launched in parallel}
% !!!!!!!!!!!Spyder details!!!!!!!!!!!!!!!!!

\section{Introduction}
\label{sec:intro}

In recent years, deep learning has shown great potential in different domains such as compression \cite{xu2022multi, schiopu2019deep}, 6D pose estimation \cite{ke2020gsnet, lyu2022mono6d} and semantic segmentation \cite{zhao2021point, qi2017pointnet++}. While most papers focus on pure performance and are able to outperform traditional methods significantly, less attention has been given to practical features. One such feature is scalability. 

Scalability is a broad term that can be applied on different aspects of deep learning, leading to different subdomains. In \cite{tan2020efficientdet, lin2017runtime}, techniques are proposed allowing the selection of the model complexity at runtime depending on the available computing resources, thus achieving complexity scalability. In \cite{royen2021masklayer}, a novel layer, called MaskLayer, is proposed that provides quality scalability with applications presented in compression and semantic hashing. The domain of resolution scalability allows operating at different resolutions, dependent on the application or available data. It has proven to be an important feature in traditional compression algorithms \cite{denis2010scalable, taquet2012hierarchical, khalil2018scalable} and, more recently, in a point cloud geometry codec \cite{guarda2020deep}. While all of these methods provide various scalability functionalities, full-resolution point cloud data is required to be available at the start of inference as input for these methods. Consequently, existing methods are not able to handle varying spatial resolutions of the input point cloud. 

%  There is finding a scalable representation and there is processing the resulting scalable representation \cite{mohan2019optic} is able to cope with a scalable representation for 2D images
% spatial scalability: The sem seg paper
% \cite{scardapane2020should}: early exit 
% Methods such as .. attempt to perform early predictions such that parts of the network can be omited. 
% \cite{li2015convolutional, sun2016pronet} suggested the usage of cascaded networks that are able to perform an early prediction and thus do not require the whole network to be executed.
% Why Scalable
% \begin{itemize}
%     \item Processing large inputs, inevitably leads to a large latency, e.g. PT for ... points takes ... ms. For many applications such as autonomous driving, this is too long to have a first estimate.
%     \item In some applications, the data is retrieved in a scalable manner. E.g. compression where a coarse estimation is refined. New scanners allow to retrieve data in a scalable manner. Scalable approaches allow to process the data immediately while data gets available and refining while more data gets available
% \end{itemize}
% Additionaly, the recent advent of scalable 3D acquisition devices \cite{vanderTempel2023Low, voxelsensors} enables the acquisition of a 3D point cloud in a scalable manner. More specifically, the obtained point cloud increases progressively in density.
In addition, existing methods are not able to progressively process additional points in the input point cloud as they become available over time. The recent advent of scalable 3D acquisition devices \cite{vanderTempel2023Low, voxelsensors} enables the acquisition of point clouds of which the densities increase progressively over time. Such resolution-scalable 3D scanning devices generate a low spatial resolution of the scene with extremely small latency, and progressively increase the resolution of the acquired point cloud over time. An important advantage of this new 3D scanning paradigm is that it enables processing the sparse point cloud while higher resolutions are captured. Once new points are captured, the results are refined. 
%Such resolution-scalable 3D scanning devices promote the design of resolution-scalable architectures, as detailed next.

\begin{figure*}[!t]
    \centering
    \includegraphics[width=.8\textwidth]{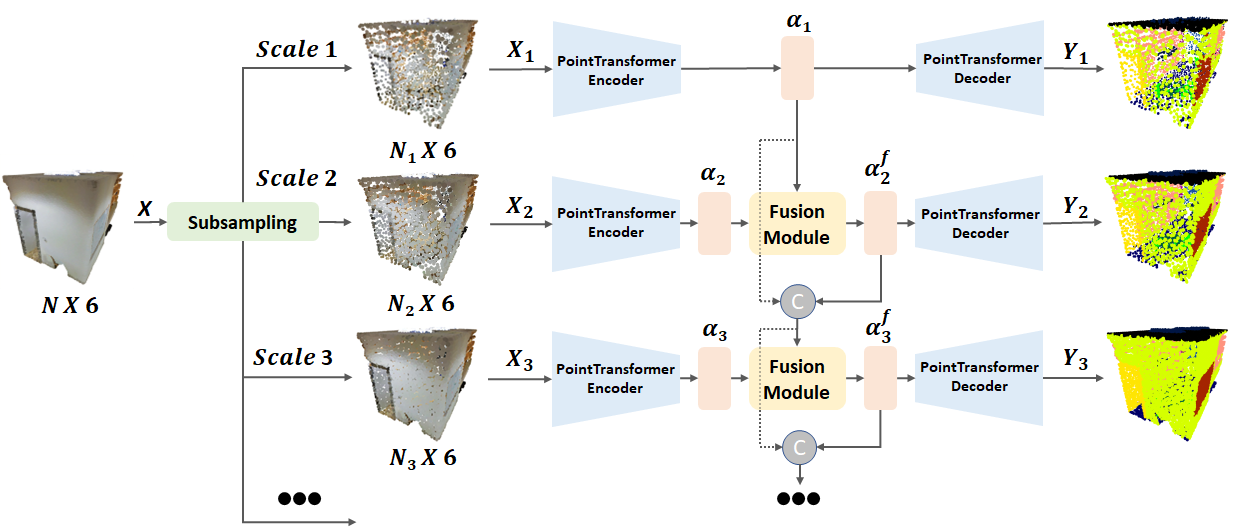}
    \caption{The RESSCAL3D architecture. The grey circle with 'C' stands for concatenation.}
    \label{fig:arch}
\end{figure*}

\begin{figure}[!t]
    \centering
    \includegraphics[width=.9\linewidth]{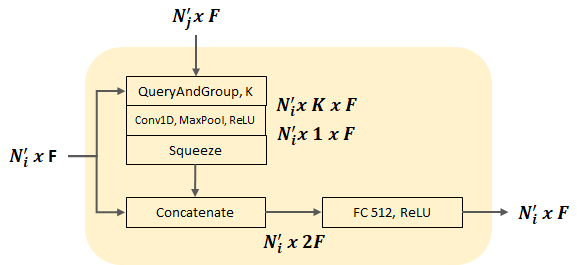}
    \caption{RESSCAL3D fusion module}
    \label{fig:arch_fusion}
\end{figure}

In this work, we propose a novel method, dubbed RESSCAL3D, that allows processing the point cloud data in a resolution-scalable manner. It allows processing low resolution 3D point clouds while higher resolutions are still being captured by the scanning device. When extra points become available, instead of restarting processing for all points, leading to large delays, the proposed method processes only the new points. To improve performance, processing of any given spatial resolution employs the information obtained at the lower spatial resolutions as prior information. This reduces the processing time. Another advantage is that early decision-making is enabled as intermediate predictions on the lower resolutions are retrieved very fast.

To evaluate the proposed architecture, semantic segmentation was chosen as target application. 3D scene understanding is of critical importance for many application domains, such as virtual reality, autonomous driving, and robotics, where timing is crucial. To this end, a fundamental component is 3D semantic segmentation \cite{qi2017pointnet++, li2018pointcnn, thomas2019kpconv, zhao2019pointweb, zhao2021point, qian2022pointnext}. We highlight PointTransformer \cite{zhao2021point} which yields state of the art results by using the Transformer architecture for this task. 

 % As a consequence, a comprehensive literature exists \cite{qi2017pointnet++, li2018pointcnn, thomas2019kpconv, zhao2019pointweb, zhao2021point, qian2022pointnext}.

 % To the best of our knowledge, no paper presinting a scalable approach for 3D semantic segmentation exists.

Summarized, our main contributions are as follows:
\begin{itemize}
    \item The first deep learning-based approach, to the best of our knowledge, that provides resolution scalable 3D semantic segmentation
    \item A fusion module that fuses features from different resolution levels
    \item An experimental analysis on S3DIS. While minimizing the cost of scalability, RESSCAL3D is 31-62\% faster than the non-scalable baseline at the highest spatial resolution. Additionally, intermediate results are generated, the fastest after only 6\% of the total inference time of the baseline.

    % \item An experimental analysis on S3DIS. While minimising cost of scalability, RESSCAL3D is not only able to yield intermediate results  \textcolor{red}{the fastest after only 6\% of the total inference time of the baseline, its total inference time exhibits a decrease of 31-62\%.}
\end{itemize}

The paper is structured as follows: \cref{sec:Proposed method} introduces the proposed approach. \cref{sec:experiments} and \cref{sec:ablation} present our experimental results and ablation study, respectively. Finally, \cref{sec:conclusion} concludes this work.
\section{Proposed method}
\label{sec:Proposed method}

% \subsection{Overview of the proposed method}
\textbf{Overview of the proposed method.}
The RESSCAL3D architecture is illustrated in \cref{fig:arch}. To retrieve the multi-resolution data, the complete input sample $\boldsymbol{X} \in \mathbb{R}^{N \times C}$, with $N$ and $C$ the number of points and channels, respectively, is subsampled in $s$ different, non-overlapping partitions. We will denote these partitions as $\boldsymbol{X}_{i} \in \mathbb{R}^{N_{i} \times C}$ with $i \in [1, s]$ and $N_{1}<...<N_{s}<N$. The employed subsampling method is described in \cref{sec:experiments}. 

Firstly, the partition with the lowest resolution, $\boldsymbol{X}_{1}$, is processed by a PointTransformer \cite{zhao2021point}, resulting in a prediction $\boldsymbol{Y}_{1} \in \mathbb{R}^{N_{1}}$. As $N_{1} << N$, the computational complexity of this first scale is low and a fast prediction can be obtained. The second scale receives as input $\boldsymbol{X}_{2}$, which is processed by another PointTransformer encoder to produce the features $\boldsymbol{\alpha}_{2} \in \mathbb{R}^{N'_{2} \times F}$, with $N'_{2}$ and $F$ the number of subsampled points by the encoder and features, respectively. In order to improve performance, those features are fused with the already computed features of lower scales by a fusion module. The resulting multi-resolution features $\boldsymbol{\alpha}_{2}^{f}$ are employed by the decoder to obtain $\boldsymbol{Y}_{2}$. At higher scales, the input of the fusion module is the concatenation of the fused features of previous scales. Once all scales are processed, $\boldsymbol{Y} = \{\boldsymbol{Y}_{1}, \boldsymbol{Y}_{2}, …, \boldsymbol{Y}_{s}\} \in \mathbb{R}^{N}$ is obtained. 

Regarding computational complexity, the presented approach has a benefit over handling all the data at once. Since PointTransformer uses an attention mechanism that requires the computation of the K-Nearest Neighbors (KNN), the complexity of processing the input as a whole can be expressed as:
% \begin{equation}
% \label{eq:comp_baseline}
$O(N^{2})=O((N_{1}+ ...+N_{s})^{2})= O(N_{1}^{2}+...+N_{s}^{2}+2\sum_{k=1}^{s}\sum_{p=1, p\neq k}^{s}N_{k}N_{p}),$
% \end{equation}
with $N=N_{1}+...+N_{s}$, and considering s scales. 

With RESSCAL3D, the attention mechanism is applied in parallel on the partitions, leading to complexity of order $O(N_{1}^{2}+...+N_{s}^{2}).$
%\begin{equation}
%\label{eq:comp_resscal3d}
 %   .
%\end{equation}
Compared to the non-scalable approach, RESSCAL3D substantially lowers complexity with a factor proportional to:
\begin{equation}
\label{eq:gain}
    \sum_{k=1}^{s}\sum_{p=1, p\neq k}^{s}N_{k}N_{p}.
\end{equation}

With a larger $s$, this effect becomes more pronounced as the partition sizes become smaller. It should be noted that sequential processing of scales also brings some computational redundancy, though the KNN is the most computational expensive operation. For large $N$ and a large amount of scales, the effect of the double product elimination becomes significant. Experimental validation of the introduced concepts is further reported in \cref{sec:experiments}.

\noindent\textbf{Fusion Module.}
Let $\boldsymbol{\alpha}^{c}_{i-1} \in \mathbb{R}^{N'_{j} \times F}$ be the concatenated features from the lower scales with $N'_j$ the number of concatenated points in feature space. Given $\boldsymbol{\alpha}^{c}_{i-1}$ and $\boldsymbol{\alpha}_{i}$, the fusion module combines the multi-scale information into a single feature matrix which is used for decoding. The fusion architecture is depicted in \cref{fig:arch_fusion}.
% Given the concatenated features from the lower scales, $\boldsymbol{\alpha}^{c}_{i-1} \in \mathbb{R}^{N'_{j} \times F}$ with $N'_j$ the number of concatenated points, and the features obtained by the encoder at the current scale, $\boldsymbol{\alpha}_{i}$, the fusion module combines the multi-scale information into a single feature matrix which is used for decoding. The fusion architecture is depicted in \cref{fig:arch_fusion}.
In more detail, the fusion module firstly retrieves the relevant features from the previous scales. This is done with a KNN algorithm on the points associated to the features in $\boldsymbol{\alpha}_{i}$. In other words, for each feature vector in $\boldsymbol{\alpha}_{i}$, the features of the K nearest neighbors in $\boldsymbol{\alpha}^{c}_{i-1}$ are utilized. As these features are originating from different resolution scales, the acquired feature matrices contain multi-resolution information. In a next step, these neighborhoods are processed by a Conv1D, followed by a MaxPool layer. After concatenation with the original scale features, $\boldsymbol{\alpha}_{i}$, a fully-connected layer encodes the information back to the original feature size.

% To do so, the fusion module firstly retrieves correspondences between the two feature sets. Since each row in the feature matrices are associated to specific points, these correspondences can be found with a K-Nearest Neighbor (KNN) algorithm. While this algorithm can be time-consuming, it is applied on the feature space which has a reduced dimensionality in the number of points, $N_j << N$. As a consequence the added time complexity is only minimal. Using the KNN, we retrieve $K$ feature vectors per point from the lower scales per feature vector in $\boldsymbol{\alpha}_{i}$. These feature

\noindent\textbf{Training.}
RESSCAL3D is trained scale by scale. All weights from previous scales are freezed while training an extra scale and the loss-function is computed only on the results from the current scale. This allows the PointTransformer backbone to achieve maximal results for each resolution.

\section{Experiments}
\label{sec:experiments}

\begin{figure}[!t]
    \centering
    \includegraphics[width=.8\linewidth]{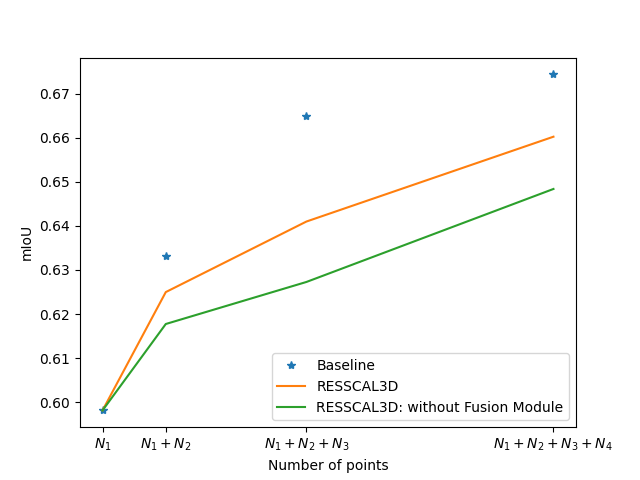}
    \caption{Ablation study and comparison of the scalable RESSCAL3D with the non-scalable baseline}%Comparison in mIoU of RESSCAL3D with the non-scalable baseline and the ablation of the Fusion Module.}
    \label{fig:mIoU_results}
\end{figure}

% \subsection{Dataset and Evaluation metrics}
\textbf{Dataset and Evaluation metrics.}
The Stanford 3D Indoor Scene dataset (S3DIS) \cite{armeni20163d} consists of 6 large-scale indoor areas with in total 271 rooms. Each point has been annotated with one of the 13 semantic categories. Area-5 has been captured in a different building than the other areas and is therefore often selected as test set \cite{zhao2021point, zhao2019pointweb, qian2022pointnext}. As evaluation metrics, the mean intersection over union (mIoU), mean accuracy (mAcc) and overall accuracy (oAcc) are being used. All presented results are averaged over the Area-5 testset.

% \subsection{Implementation details}
% \label{subsec:impl_details}
\noindent\textbf{Implementation details.} 
PointTransformer \cite{zhao2021point} has been selected as backbone architecture as it achieves state of the art performance and has official, publicly available code. Each scale was trained for 34 epochs with a batch size of 4. Other training and network parameters are the same as in \cite{zhao2021point}. Each input point is represented by a 6-dimensional vector: $xyzrgb$. To obtain the multi-resolution data, $\boldsymbol{X}$ is voxelized $s$ times with $s$ different voxel sizes. Subsequently, one point per voxel is randomly selected while making sure a point is not present in multiple partitions. More specifically, we have opted to employ 4 scales with voxel sizes [0.16, 0.12, 0.08, 0.06].

% \textcolor{red}{While the subsampling depends on the practical applications of RESSCAL3D, we simulate a possible subsampling.} To do so, $\boldsymbol{X}$ is voxelized $s$ times with $s$ different voxel sizes.

\begin{figure}[!b]
    \centering
    \includegraphics[width=.8\linewidth]{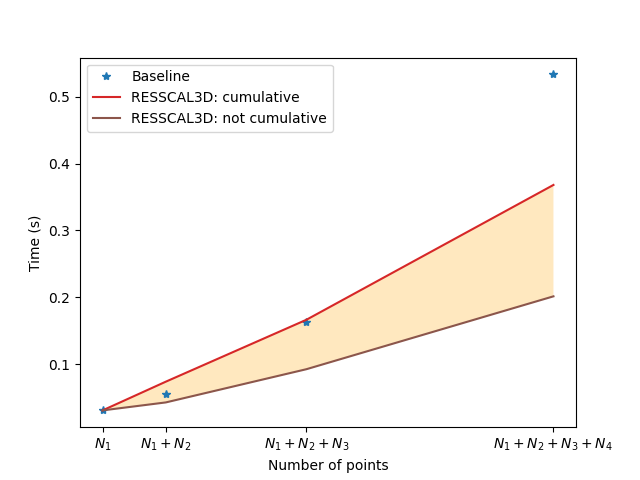}
    \caption{Comparison of RESSCAL3D with the non-scalable baseline in inference time.  The actual  inference latency is bounded to the yellow zone. The displayed non-scalable baseline timing results are not cumulative.}
    % \caption{Comparison of RESSCAL3D with the non-scalable baseline in inference time. As RESSCAL3D's scalable approach allows to process the data during the acquisition, the specifically introduced latency by RESSCAL3D depends on the data acquistion time. Nevertheless, it can be upper bounded by the edge case where the all data is immediately available and all latency introduced is solely by RESSCAL3D (red line). The lower bound is the case where all computations are done before the next batch of data becomes available (brown line). Thus, the real inference latency is situated in the orange area. The displayed non-scalable baseline timing results are not cumulative.}
    \label{fig:time_results}
\end{figure}

\begin{figure*}[t]
\centering
\captionsetup{justification=centering}
{\includegraphics[width=0.24\textwidth]{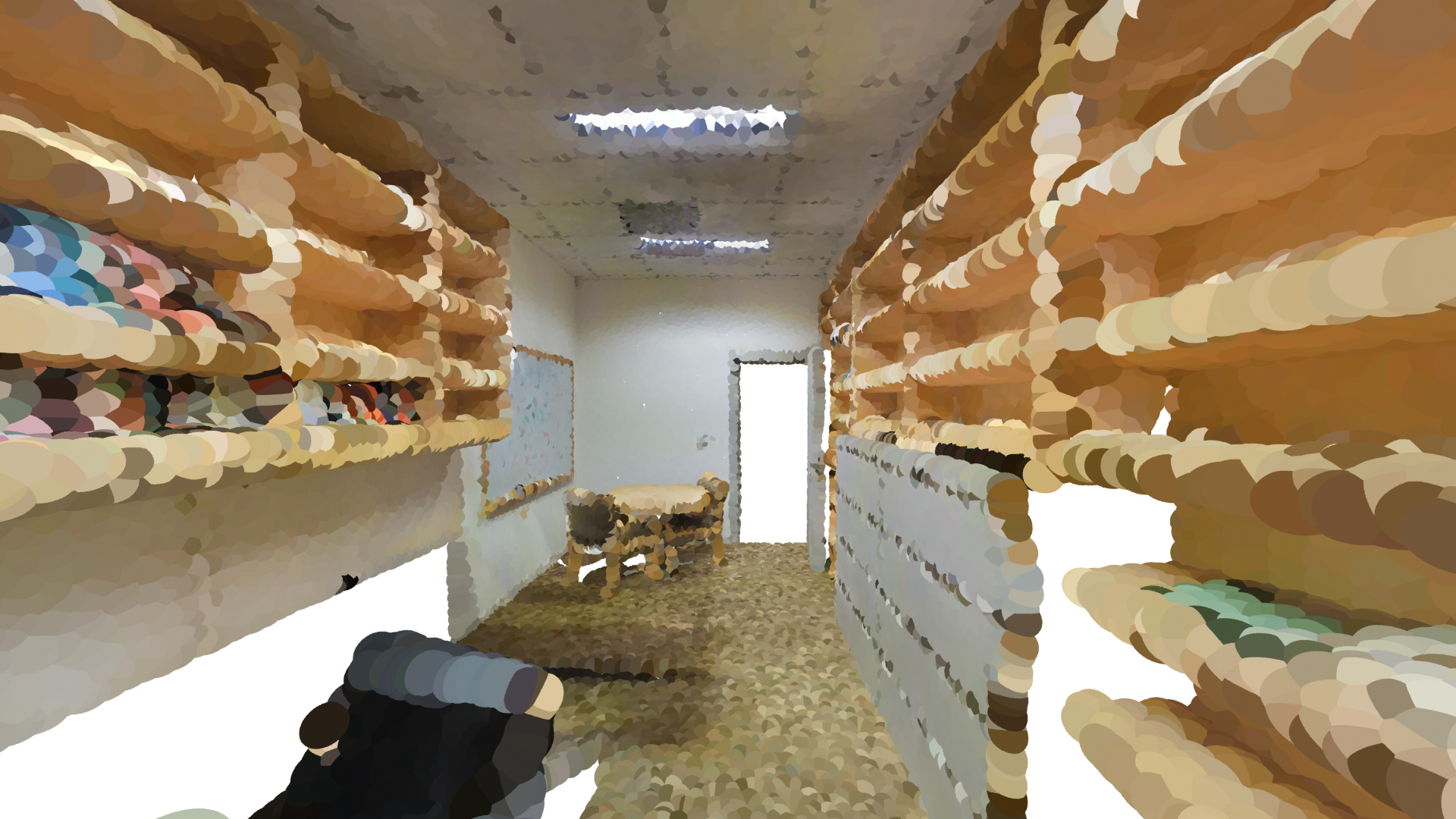}}
{\includegraphics[width=0.24\textwidth]{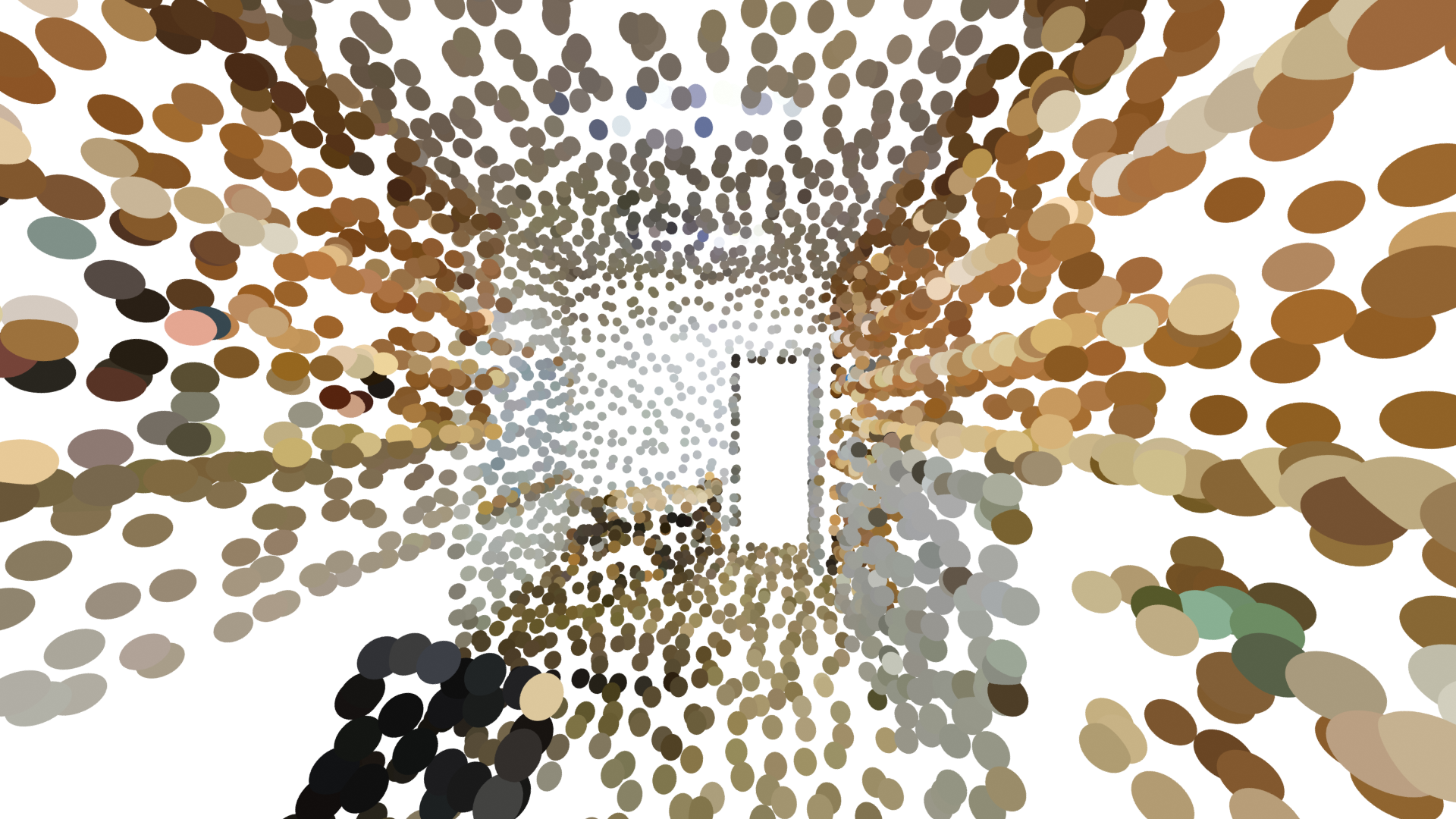}}
{\includegraphics[width=0.24\textwidth]{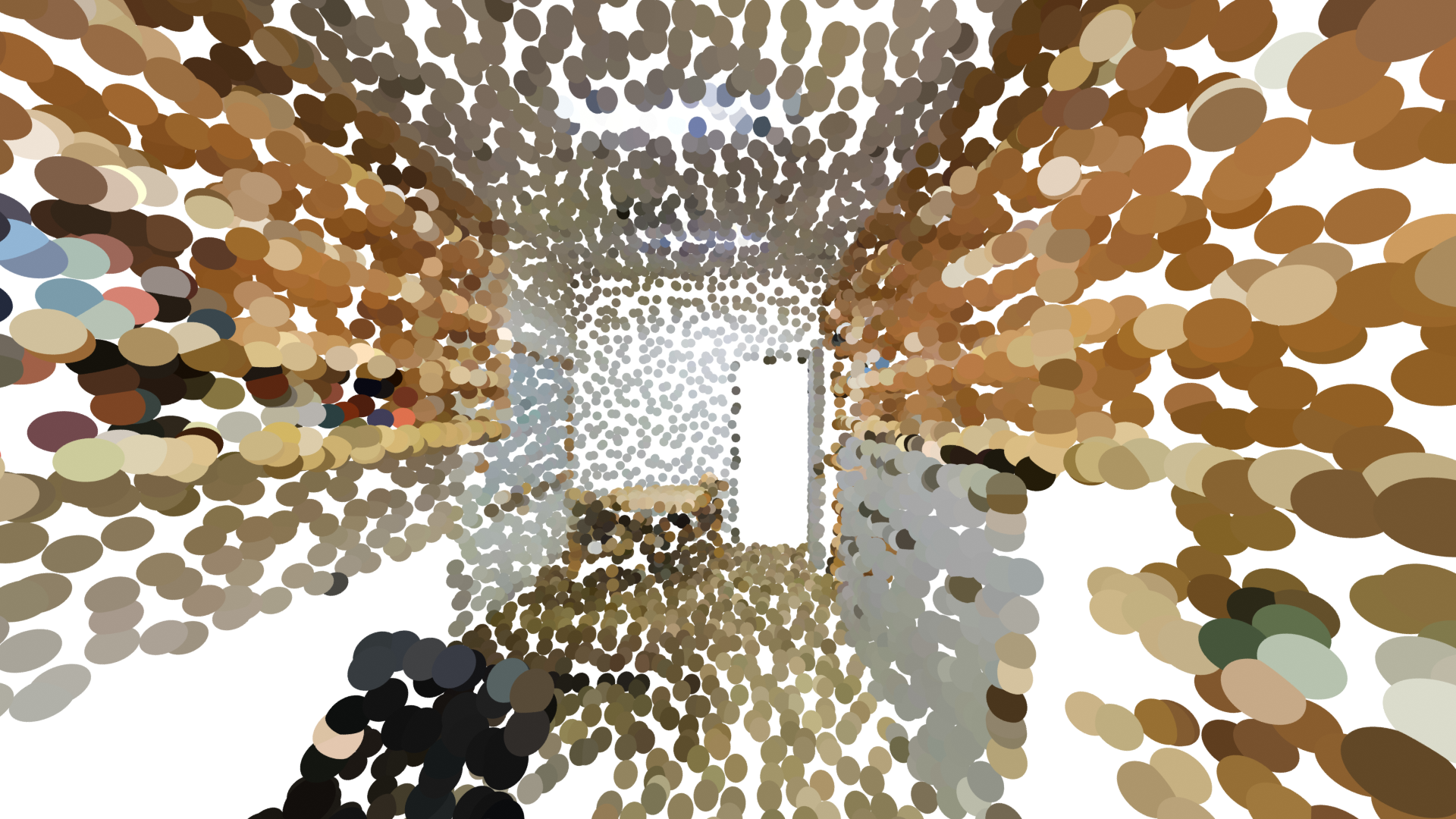}}
{\includegraphics[width=0.24\textwidth]{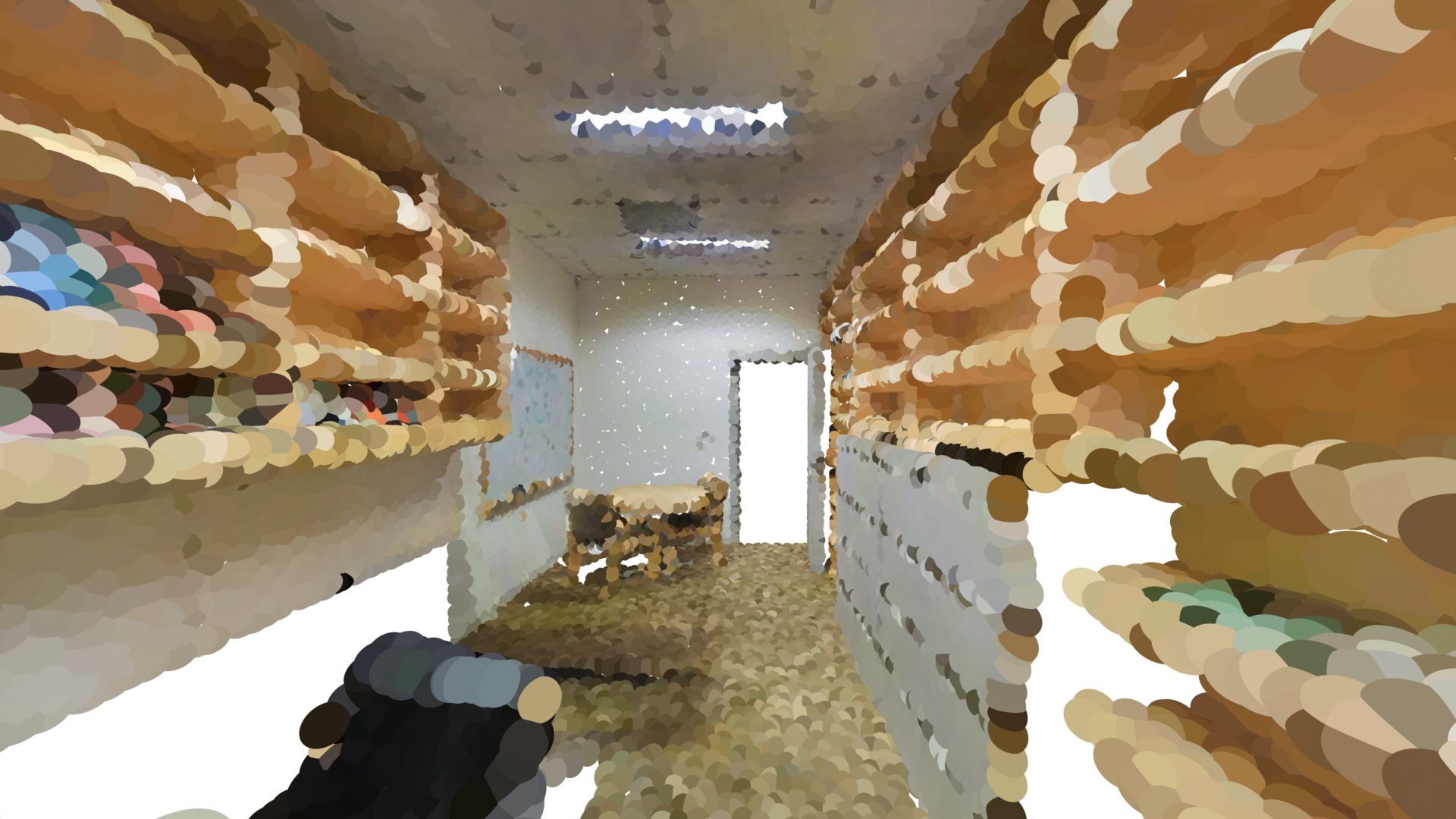}}
% \subcaptionbox*{(d)\\Scale 2\\Time: \textcolor{red}{???} s}{\includegraphics[width=0.19\textwidth]{figs/arch/resscal_fusion_arch.png}} 
\subcaptionbox*{(a)\\Ground truth}{\includegraphics[width=0.24\textwidth]{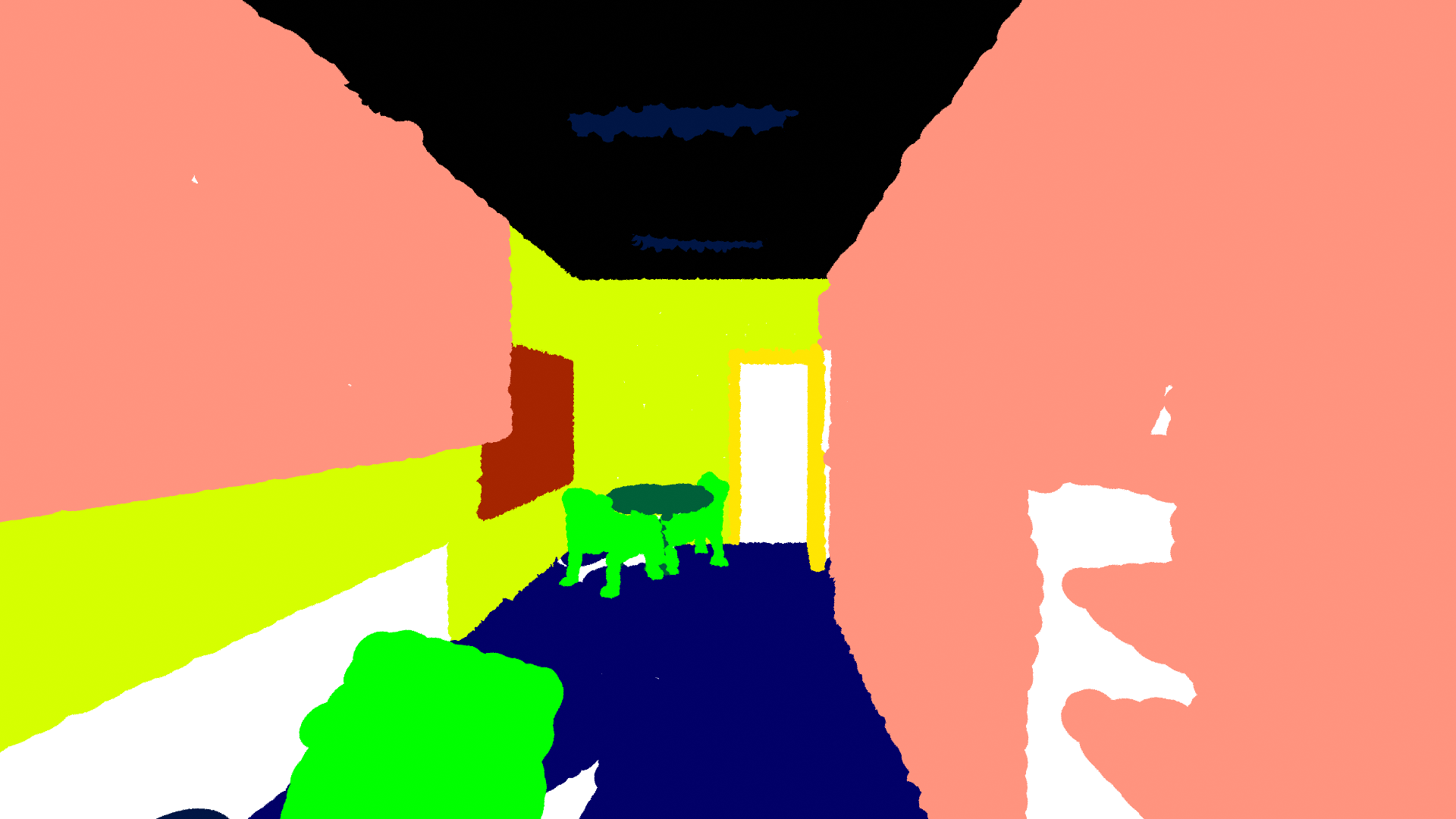}}
\subcaptionbox*{(b)\\Scale 0\\Time: 31.1 ms}{\includegraphics[width=0.24\textwidth]{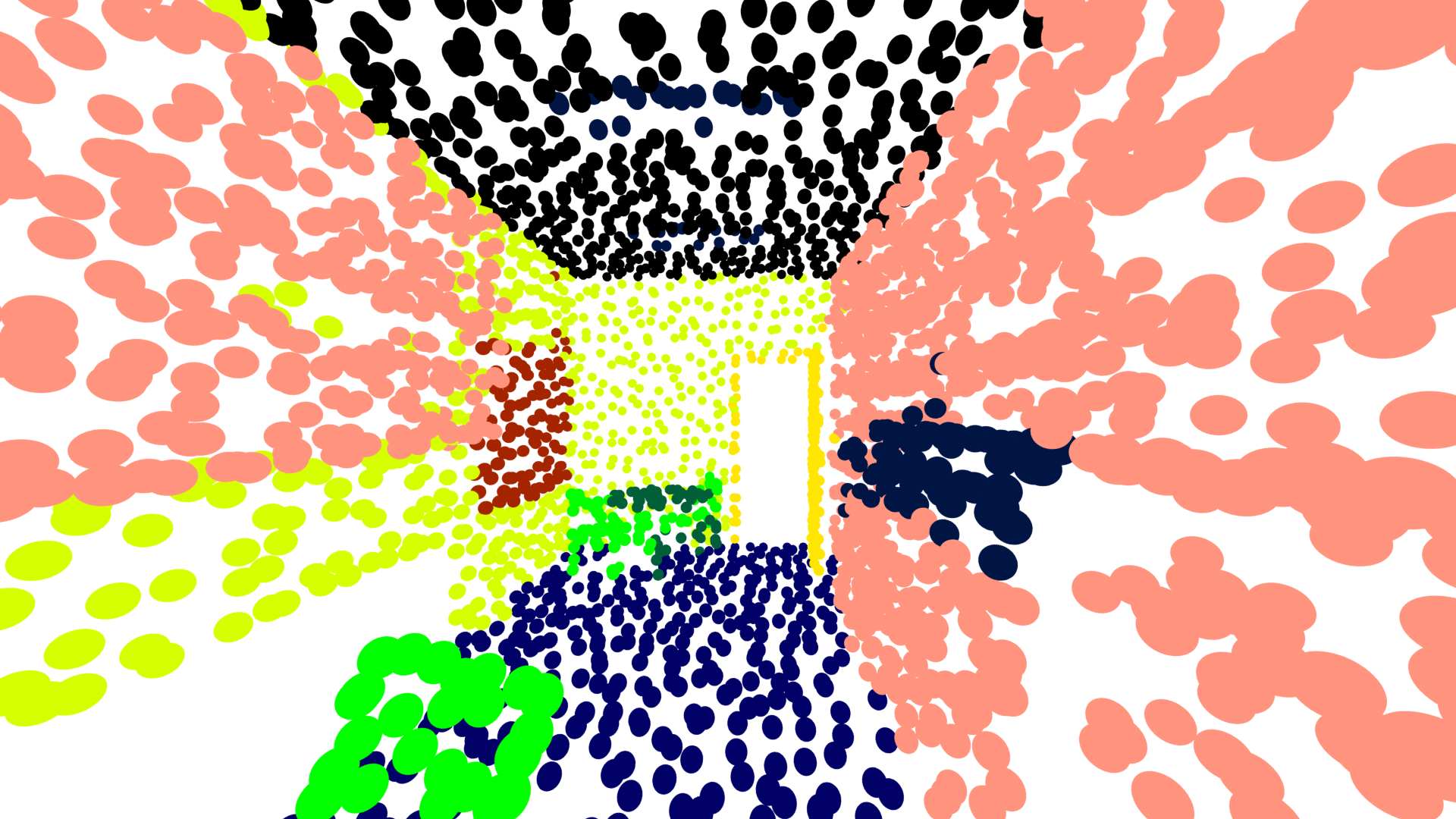}}
\subcaptionbox*{(c)\\Scale 1\\Time: 42.9 ms}{\includegraphics[width=0.24\textwidth]{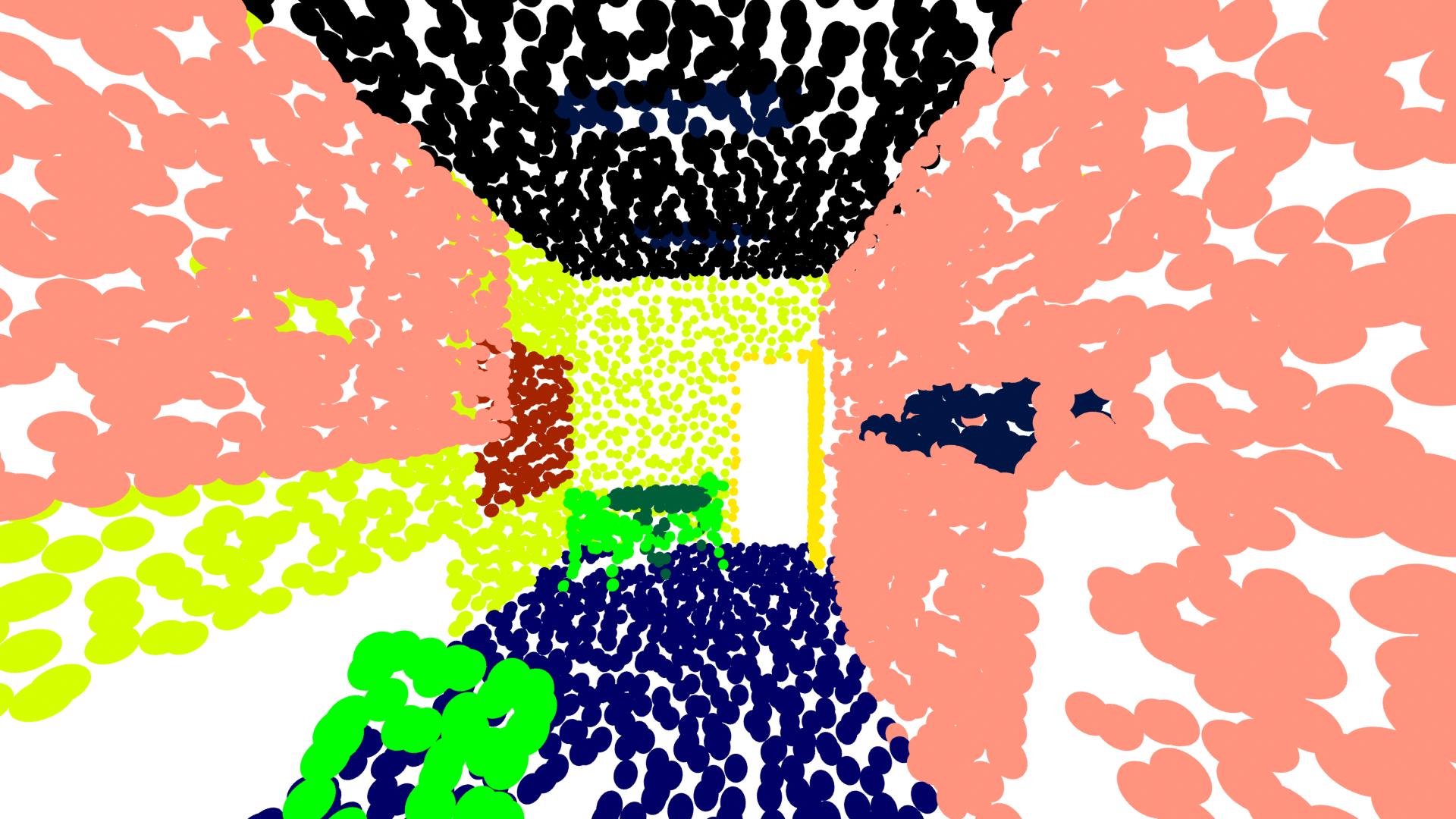}}
\subcaptionbox*{(d) \\Scale 3\\Time: 201 ms}{\includegraphics[width=0.24\textwidth]{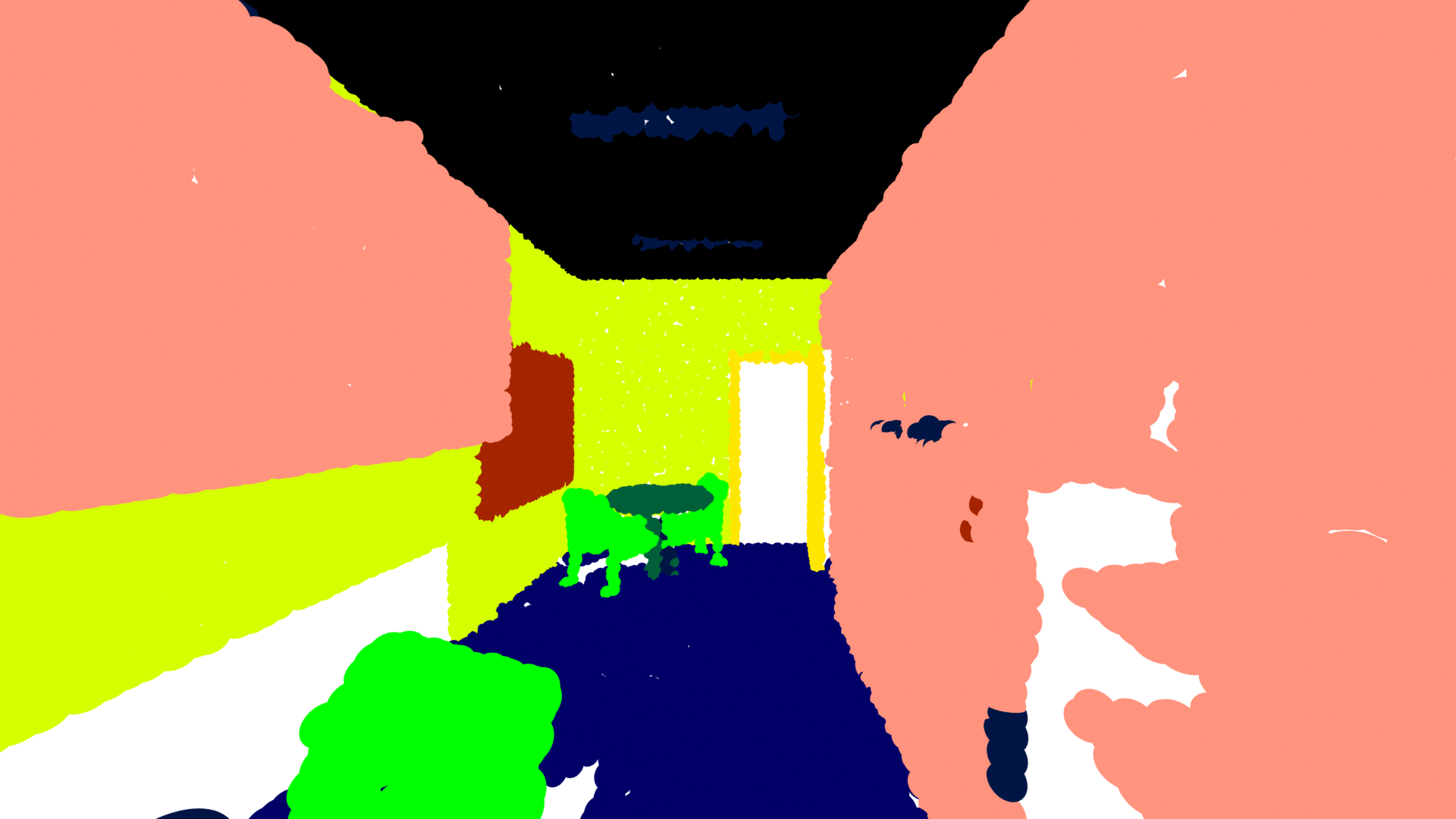}}
\captionsetup{justification=justified}
\caption{Visualization of S3DIS results for RESSCAL3D. The input data and semantic prediction are visualized on the top and bottom row, respectively. Non-cumulative time is used.}
\label{fig:visuals}
\end{figure*}

% \subsection{S3DIS semantic segmentation}
\noindent\textbf{S3DIS semantic segmentation.}
RESSCAL3D is, to the best of our knowledge, the first method that performs resolution-scalable 3D semantic segmentation of point clouds. Consequently, no quantitative or qualitative comparison with existing methods can be made. Nevertheless, in order to characterize its performance and inference time, we compare the proposed method with the non-scalable baseline, which employs the same semantic segmentation backbone for the different scales. Our scalable approach processes the additional points at each scale, using side information from the previous scales, the baseline processes the whole point cloud at that scale. Thus, the latter can only be launched when all data is available and does not process data in a progressive manner. Also, no intermediate results are obtained. 

The results in terms of mIoU of our scalable approach, with and without the fusion module, and the non-scalable baseline are shown in \cref{fig:mIoU_results}. At the first scale, all methods operate in an identical manner and thus, achieve equal performance. At higher scales, using the proposed fusion module reduces the performance gap between the scalable approach and non-scalable baseline. At the highest scale, the performance gap in mIoU is only 2.1\% of the total performance. Although the proposed method is not able to achieve the same performance as the baseline at the highest scale, the resulting difference is deemed small.%It can be seen as the cost of scalability as RESSCAL3D does not process the whole points at once. 

On the other hand, the scalable approach presents an important advantage in inference time. In \cref{fig:time_results}, the inference time can be compared. Important to note is that the latency invoked by RESSCAL3D depends on the data availability. Since the scalable approach can start processing lower resolutions while higher resolutions are being acquired, it utilizes the otherwise lost acquisition time, while the non-scalable baseline can only start once all point cloud data is available. Therefore we have opted to present the upper and lower bounds of the induced latency by RESSCAL3D. When operating on the upper bound, all data is available at the start and all inference timings are cumulated. For the latter, the processing of the previous scale is finished before the point cloud data for the current resolution becomes available. Therefore, the latency introduced by RESSCAL3D will be in the yellow zone (see \cref{fig:time_results}) and is mainly lower than the baseline. Important to note is that even if operating on the upper bound,  RESSCAL3D is able to attain a 31\% decrease in inference time at the highest scale with respect to the baseline. The main reason is the reduced complexity of the attention modules as explained in \cref{sec:Proposed method}. In the lower bound case, RESSCAL3D achieves an impressive 61\% decrease in inference time. 
% \cref{eq:gain} shows that when operating with higher number of points, this gain will become even more pronounced.
When operating with higher number of points, the gain will become even more pronounced (\cref{eq:gain}).

Qualitative results are presented in \cref{fig:visuals}. Overall, one can notice very accurate segmentation, with some errors on the lower scales corrected at the higher scales. An example is the erroneous dark segmentation on the bookcase on the right.

% \begin{table}[!t]
% \resizebox{\columnwidth}{!}{
% \begin{tabular}{|c|c|c|c|c|c|c|}
% \hline
% \multirow{2}{*}{Scale} & \multirow{2}{*}{Method} & \multicolumn{3}{c|}{Performance} & \multirow{2}{*}{Time (ms)} \\ \cline{3-5} 
% &                        &  oAcc & mAcc & mIoU &  \\ \hline
% \multirow{3}{*}{0} & Baseline                &  85.7 & 67.9 & 59.8 & 31.1 \\ \cdashline{2-6}
% & Ours - no data latency    &  \multirow{2}{*}{85.7} & \multirow{2}{*}{67.9} & \multirow{2}{*}{59.8} & 31.1 \\
% & Ours - large data latency    &   &  &  & 31.1\\ \hline
% \multirow{3}{*}{1} & Baseline                &  87.3 & 71.1 & 63.3 & 55.6 \\ \cdashline{2-6}
% & Ours - no data latency    &  \multirow{2}{*}{87.0} & \multirow{2}{*}{70.0} & \multirow{2}{*}{62.5} & 73.9 \\
% & Ours - large data latency    &   &  &  & 42.9\\ \hline
% \multirow{3}{*}{3} & Baseline                &  88.5 & 73.5 & 66.5 & 164 \\ \cdashline{2-6}
% & Ours - no data latency    &  \multirow{2}{*}{87.6} & \multirow{2}{*}{71.2} & \multirow{2}{*}{64.1} & 167 \\
% & Ours - large data latency    &   &  &  & 92.6\\ \hline
% \multirow{3}{*}{4} & Baseline                &  88.9 & 74.1 & 67.4 & 533 \\ \cdashline{2-6}
% & Ours - no data latency    &  \multirow{2}{*}{88.5} & \multirow{2}{*}{73.0} & \multirow{2}{*}{66.0} & 368 \\
% & Ours - large data latency    &   &  &  & 201\\ \hline
% \end{tabular}
% }
% \caption{Quantitative results}
% \label{tab:comp_gsnet50}
% \end{table}

\begin{table}[!t]
\resizebox{\columnwidth}{!}{
\begin{tabular}{|c|c|c|c|c|c|c|}
\hline
\multirow{2}{*}{Scale} & \multirow{2}{*}{Method} & \multicolumn{3}{c|}{Performance} & \multirow{2}{*}{Time (ms)} \\ \cline{3-5} 
&                        &  oAcc & mAcc & mIoU &  \\ \hline
\multirow{2}{*}{0} & Without Fusion               &  85.7 & 67.9 & 59.8 & 31.1 \\
& Fusion &  85.7 & 67.9 & 59.8 & 31.1 \\ \hline
\multirow{2}{*}{1} & Without Fusion                &  86.5 & 69.6 & 61.8 & \textbf{73.8} \\
& Fusion   &  \textbf{87.0} & \textbf{70.0} & \textbf{62.5} & 73.9 \\ \hline
\multirow{2}{*}{3} & Without Fusion                &  87.0 & 70.5 & 62.7 & 167 \\ 
& Fusion    &  \textbf{87.6} & \textbf{71.2} & \textbf{64.1} & 167 \\ \hline
\multirow{2}{*}{4} & Without Fusion                &  87.8 & 72.4 & 64.8 & 368 \\
& Fusion    &  \textbf{88.5} & \textbf{73.0} & \textbf{66.0} & 368 \\ \hline
\end{tabular}
}
\caption{Ablation of fusion module. Cumulative time is used.}
\label{tab:ablation}
\end{table}

% \begin{table}
% \centering
% \begin{tabular}{|c|c|c|c|} 
% \hline
% \multirow{2}{*}{Method} & \multicolumn{2}{c|}{R} & T     \\ 
% \cline{2-4}
%                         & $acc(\frac{\pi}{6})$ & $Mederr$ & $ARED$  \\ 
% \hline
% Mono6D-rgb                & 93.47\%    &  3.24                & 6.57\%      \\ 
% \hline
% Mono6D                    & \textbf{93.80\%}    &  \textbf{3.00}                & \textbf{4.76\%}      \\
% \hline
% \end{tabular}
% \caption{Effect of K in fusion module}
% \label{tab:ablation}
% \end{table}
\section{Ablation Study}
\label{sec:ablation}

In this section, the effect and value of our fusion module is analysed. The removal of the fusion module leads to the loss of multi-resolution processing and scales which are processed independently. In \cref{fig:mIoU_results} and \cref{tab:ablation} is shown that employing the fusion module consistently leads to better results. The added inference time is negligible.
\section{Conclusion}
\label{sec:conclusion}

In this paper, we propose RESSCAL3D, a novel architecture allowing resolution scalable 3D semantic segmentation of point clouds. The experiments show that our scale-by-scale approach allows significantly faster inference while maintaining a limited impact on performance relative to the non-scalable baseline.

% \section{Acknowledgement}
% \label{sec:ack}

% This is acknowledgement

% Below is an example of how to insert images. Delete the ``\vspace'' line,
% uncomment the preceding line ``\centerline...'' and replace ``imageX.ps''
% with a suitable PostScript file name.
% -------------------------------------------------------------------------
% \begin{figure}[htb]

% \begin{minipage}[b]{1.0\linewidth}
%   \centering
%   \centerline{\includegraphics[width=8.5cm]{image1}}
% %  \vspace{2.0cm}
%   \centerline{(a) Result 1}\medskip
% \end{minipage}
% %
% \begin{minipage}[b]{.48\linewidth}
%   \centering
%   \centerline{\includegraphics[width=4.0cm]{image3}}
% %  \vspace{1.5cm}
%   \centerline{(b) Results 3}\medskip
% \end{minipage}
% \hfill
% \begin{minipage}[b]{0.48\linewidth}
%   \centering
%   \centerline{\includegraphics[width=4.0cm]{image4}}
% %  \vspace{1.5cm}
%   \centerline{(c) Result 4}\medskip
% \end{minipage}
% %
% \caption{Example of placing a figure with experimental results.}
% \label{fig:res}
% %
% \end{figure}

%\vfill
%\pagebreak

\vfill\pagebreak

% \section{REFERENCES}
% \label{sec:refs}

% List and number all bibliographical references at the end of the
% paper. The references can be numbered in alphabetic order or in
% order of appearance in the document. When referring to them in
% the text, type the corresponding reference number in square
% brackets as shown at the end of this sentence \cite{C2}. An
% additional final page (the fifth page, in most cases) is
% allowed, but must contain only references to the prior
% literature.

\bibliographystyle{IEEEbib.sty}
\bibliography{refs}

\end{document}